\documentclass{article}

\usepackage{graphicx}
\usepackage{color}
\usepackage{subfigure}
\usepackage{booktabs} 
\usepackage{listings}
\usepackage{multicol}
\usepackage{xcolor}

\lstdefinelanguage{JavaScript}{
  morekeywords=[1]{break, continue, delete, else, function, if, in,
    new, this, typeof, var, void, while, with, constructor},
  morekeywords=[2]{false, null, true, boolean, number, undefined,
    Array, Boolean, Date, Math, Number, String, Object},
  morekeywords=[3]{eval, parseInt, parseFloat, escape, unescape},
  sensitive,
  morecomment=[s]{/*}{*/},
  morecomment=[l]{//},
  morecomment=[s]{/**}{*/}, 
  morestring=[b]{'},
  morestring=[b]{"},
}[keywords, comments, strings]

\lstalias[]{ES6}[ECMAScript2015]{JavaScript}
\lstdefinelanguage[ECMAScript2015]{JavaScript}[]{JavaScript}{
  morekeywords=[1]{async, case, catch, class, const, default, do,
    enum, export, extends, finally, from, implements, import, instanceof,
    let, static, super, switch, throw, try},
  morekeywords=[4]{await, return, for},
  morestring=[b]{`} 
}

\lstdefinelanguage[distmljsSample]{JavaScript}[ECMAScript2015]{JavaScript}{
  morekeywords=[2]{MLPModel, K, nn, core, Layer, layers, Promise, VariableResolvable, functions, images, labels, trainLoader, optimizer},
  morekeywords=[5]{c, relu, Linear, tidy, Variable, to, softmaxCrossEntropy, zeroGrad, backward, step},
}

\definecolor{basecolor}{HTML}{081884}
\definecolor{identifiercolor}{HTML}{081884}
\definecolor{commentcolor}{HTML}{028002}
\definecolor{strcolor}{HTML}{a6201f}
\definecolor{keywordcolor1}{HTML}{1010ff}
\definecolor{keywordcolor2}{HTML}{2d839e}
\definecolor{keywordcolor4}{HTML}{af01db}
\definecolor{keywordcolor5}{HTML}{795e27}

\usepackage[hyphens]{url}
\usepackage[driverfallback=dvipdfm]{hyperref}

\usepackage[accepted]{mlsys2024}

\mlsystitlerunning{DistML.js: Installation-free Distributed Deep Learning Framework for Web Browsers}

\begin{document}

\twocolumn[
\mlsystitle{DistML.js: Installation-free Distributed Deep Learning Framework for Web Browsers}




\begin{mlsysauthorlist}
  \mlsysauthor{Masatoshi Hidaka}{mil}
  \mlsysauthor{Tomohiro Hashimoto}{mil}
  \mlsysauthor{Yuto Nishizawa}{mil}
  \mlsysauthor{Tatsuya Harada}{mil,riken}
  \end{mlsysauthorlist}

  \mlsysaffiliation{mil}{The University of Tokyo, Japan}
  \mlsysaffiliation{riken}{RIKEN AIP, Japan}

  \mlsyscorrespondingauthor{Tatsuya Harada}{harada@mi.t.u-tokyo.ac.jp}

\mlsyskeywords{Machine Learning, MLSys}

\vskip 0.3in

\begin{abstract}
We present ``DistML.js'', a library designed for training and inference of machine learning models within web browsers. Not only does DistML.js facilitate model training on local devices, but it also supports distributed learning through communication with servers. Its design and define-by-run API for deep learning model construction resemble PyTorch, thereby reducing the learning curve for prototyping. Matrix computations involved in model training and inference are executed on the backend utilizing WebGL, enabling high-speed calculations. We provide a comprehensive explanation of DistML.js's design, API, and implementation, alongside practical applications including data parallelism in learning. The source code is publicly available at \url{https://github.com/mil-tokyo/distmljs}.
\end{abstract}
]



\printAffiliationsAndNotice{} 

\section{Introduction}
\label{introduction}

The rapid advancement of machine learning technology continues to yield increasingly powerful models, profoundly impacting various domains. Training deep learning models demands high-performance computing resources and vast amounts of data. In contemporary training workflows, high-end servers equipped with GPUs predominantly handle computational tasks, while pre-collected data is stored on servers and accessed during training. Conversely, it's also feasible to distribute machine learning training across decentralized devices such as PCs and smartphones. Regarding computing resources, research has demonstrated the potential of volunteer computing, harnessing numerous devices to achieve computational performance surpassing supercomputers~\cite{zimmerman-2020}. In terms of data, federated learning proposes techniques to utilize data from mobile devices while preserving privacy~\cite{mcmahan-2017}. Hence, leveraging distributed devices proves advantageous for both data utilization and computational powers in machine learning.

In pursuit of applying volunteer computing to machine learning training, we propose ``DistML.js,'' a JavaScript-based machine learning library, as a novel distributed training platform. DistML.js not only facilitates training and inference of deep models on local devices but also offers training capabilities via communication with servers, tailored for distributed learning scenarios. These functionalities operate within web browsers, ensuring compatibility across a wide range of devices regardless of their type or operating system. Our contributions in this paper are summarized as follows.

\begin{itemize}
  \item A novel JavaScript machine learning library, ``DistML.js,'' has been developed and published as an open-source software. We demonstrate the functionality of training models for image recognition tasks while utilizing edge devices as hardware resources.
  \item We presented an implementation example of distributed training using data parallelism, showing accelerated training as the number of participating devices increases.
\end{itemize}

\section{Related Work}
\label{related-works}
\subsection{Platforms and libraries for machine learning}

The predominant programming languages for utilizing machine learning techniques are typically limited to Python, C++, and R. In each of these languages, sophisticated libraries have been made available for training and inference of machine learning models. For instance, in Python, the widely popular libraries such as Scikit-learn~\cite{pedregosa-2011}, TensorFlow~\cite{abadi-2016}, and PyTorch~\cite{paszke-2019} have been released, making the utilization of machine learning techniques more accessible to a broad user base.

On the other hand, JavaScript has gained significant attention in recent years as a programming language. It primarily executes within web browsers, allowing for cross-platform usage across various operating systems and hardware types. Additionally, technologies for high-performance computing, such as WebGL for 3D graphics rendering~\cite{webgl} and WebAssembly for fast execution of binary code~\cite{haas-2017}, have also advanced. While JavaScript is not commonly used in the modern machine learning field, existing libraries like TensorFlow.js enable the application of machine learning techniques in web browsers~\cite{smilkov-2019}. TensorFlow.js is, to our knowledge, the first JavaScript-based deep learning library available that supports not only inference but also training. The library introduced in this paper, DistML.js, also supports a training scheme. The relationship between TensorFlow.js and DistML.js is detailed in Section \ref{tfjs}.

Some JavaScript-based machine learning libraries such as ONNX.js~\cite{onnxjs}, its newer version ONNX Runtime Web~\cite{onnxruntime}, and WebDNN~\cite{hidaka-2017} supports inference, while training is unavaliable. Leveraging backends such as WebGL and WebAssembly, these libraries facilitate fast inference of machine learning models on web browsers.

\subsection{Distributed machine learning}

In the realm of machine learning training, parallel computing utilizing multiple computers is often employed~\cite{dean-2012,chilimbi-2014}. Particularly effective when dealing with vast datasets, data parallelism entails distributing data across multiple machines for processing, thus achieving acceleration of model training. Each calulator utilizes a splitted partition of the dataset to update the model parameters, sending the results to a central server. At the central server, the accumulated model updates are integrated, yielding results similar to those obtained from updating with a large batch size on a single machine. Data parallelism is the most common form of distributed deep learning and is supported by many machine learning libraries such as TensorFlow and PyTorch.

On the other hand, when dealing with massive models, model parallelism proves effective. By partitioning the model and distributing it across multiple machines, it becomes possible to train models with parameters that exceed the memory capacity of a single machine~\cite{shoeybi-2019,rasley-2020,zhao-2023}.

\subsection{Volunteer computing}

The mechanism of harnessing a vast array of interconnected devices over the Internet to execute extensive computations is referred to as ``volunteer computing.'' Volunteer computing predominantly involves individuals lending their personally owned devices as computational resources for socially impactful tasks, with SETI@Home~\cite{anderson-2002} and Folding@home~\cite{zimmerman-2020} being among the most notable examples of volunteer computing initiatives. SETI@Home aims to analyze radio signals observed by observatories to discover evidence of extraterrestrial life, while Folding@home focuses on analyzing protein folding and dynamics through molecular dynamics simulations. Folding@home, in particular, garnered widespread public attention due to pressing societal demands, attracting participation from millions of devices and achieving the unprecedented computational speed of ExaFLOPS.

These projects facilitate participation through dedicated software known as middleware. One well-known middleware is the Berkeley Open Infrastructure for Network Computing (BOINC)~\cite{anderson-2020}. Middleware is expected to consistently yield identical computational results across different operating systems or hardware configurations, a characteristic that aligns well with the cross-platform nature of web browsers. While participation in the Folding@home project commonly involves the use of dedicated clients, it was possible to participate in the Folding@home project directly through web browsers such as Google Chrome and Chromium~\cite{foldingatchrome}.

\subsection{Web browsers as a platform of distributed machine learning}

Attempts to execute distributed deep learning via volunteer computing have been documented. The following studies primarily reports the feasibility of achieving distributed computation using web browsers. In MLitB~\cite{meeds-2015}, it is demonstrated that image training tasks such as MNIST and CIFAR10 can be learned through distributed computation in the browser. Additionally, JSDoop~\cite{morell-2019} showcases the capability of learning tasks involving text prediction using RNNs through distributed computation in the browser while utilizing TensorFlow.js. However, it is reported that in supervised learning, communication becomes a bottleneck, leading to a plateau in training speed even with an increased number of distributed devices connected online.
\section{Implementations}
\label{implementation}

\begin{figure*}[t]
  \begin{lstlisting}[caption=An example of model definition in DistML.js.\vspace{0.5em},label=K_model]
import * as K from 'distmljs';

// define model by extending K.nn.core.Layer
class MLPModel extends K.nn.core.Layer {
  // trainable layers have to be stored as property
  l1: K.nn.layers.Linear;
  l2: K.nn.layers.Linear;

  constructor(inFeatures: number, hidden: number, outFeatures: number) {
    super();
    this.l1 = new K.nn.layers.Linear(inFeatures, hidden);
    this.l2 = new K.nn.layers.Linear(hidden, outFeatures);
  }

  // forward defines control flow
  async forward(inputs: K.nn.Variable[]): Promise<K.nn.Variable[]> {
    let y: K.nn.VariableResolvable = inputs[0];
    y = this.l1.c(y);
    y = K.nn.functions.relu(y);
    y = this.l2.c(y);
    return [await y];
  }
}
  \end{lstlisting}
\end{figure*}

\begin{figure*}[t]
  \begin{lstlisting}[caption=An example of training loop in DistML.js.\vspace{0.5em},label=K_train]
import * as K from 'distmljs';

for (let epoch = 0; epoch < epochs; epoch++) {
  for await (const [images, labels] of trainLoader) {
    // K.tidy: takes function, and releases tensor allocated in the function after it ends.
    // it is needed because GPU memory cannot be garbage collected.
    await K.tidy(async () => {
      // model (K.nn.core.Layer) can be called with model.c
      // model receives Variable, which can be constructed with new Variable(Tensor)
      // Tensor.to(backend) copies tensor data to another backend (CPUTensor/WebGLTensor)
      const y = await model.c(new K.nn.Variable(await images.to(backend)));
      const loss = await K.nn.functions.softmaxCrossEntropy(
        y,
        new K.nn.Variable(await labels.to(backend))
      );
      // remove gradient of Parameter in model.
      optimizer.zeroGrad();
      // backpropagation
      await loss.backward();
      // update Parameter in model using gradient.
      await optimizer.step();
      // Tensors inside them are kept after tidy.
      return [model, optimizer];
    });
  }
}
    \end{lstlisting}
\end{figure*}

This chapter elucidates the design and implementation of DistML.js. DistML.js encompasses many features related to deep learning. Ranging from low-level array operations such as efficient tensor representation and matrix computations to advanced features utilized in contemporary deep learning, including layer structures, model architectures, optimization algorithms, loss functions, activation functions and data loaders. All of these functionalities are implemented in TypeScript, a language that extends JavaScript with static typing capabilities and allows for transpilation of the source code to JavaScript.

\subsection{Overall}
The components of DistML.js and concepts of implementation are described below.

\begin{itemize}
  \item \textbf{Tensor Representation}: By utilizing the custom implemented Tensor type in DistML.js, efficient conversion from JavaScript arrays to tensors using TypedArray is achievable, facilitating rapid numerical computations.

  \item \textbf{Slicing}: The use of advanced slicing enables extraction and manipulation of elements in multidimensional tensors. This functionality not only proves beneficial for deep learning tasks but also renders the library useful for data preprocessing and other general scientific computing implementations.

  \item \textbf{PyTorch-like API}: Definitions of layers, models, and loss functions in deep learning provide an interface similar to PyTorch. This feature enables users familiar with PyTorch to seamlessly leverage DistML.js with minimized learning overhead.

  \item \textbf{Accelerated Computation}: By harnessing WebGL for matrix computations, GPU utilization becomes feasible. Consequently, even for large datasets and models, high-speed computations can be performed within the limits of available memory.

  \item \textbf{Tools for Distributed Learning}: As a distinctive feature, DistML.js offers tools tailored for distributed deep learning. The server is implemented in Python, facilitating seamless integration with existing code for data preprocessing and postprocessing. Additionally, a mechanism is in place for communication between web browsers as clients, facilitating the execution of computation and the communication of tensors. Through the serialization mechanism for tensor collections, models and data can be consolidated into a single binary data unit, enabling easy and efficient communication.
\end{itemize}

Some example codes for deep learning with DistML.js is presented in Code \ref{K_model} and \ref{K_train}. Code \ref{K_model} defines the class for the MLP model, while Code \ref{K_train} comprises the code for training the model.

\subsection{API design}

The API for deep learning provided by DistML.js aims to reduce the learning overhead by offering an interface similar to PyTorch~\cite{paszke-2019}. For instance, models are defined as classes, and the forward propagation of models is described within the forward method. Fully connected layers are represented by the ``\verb|Linear|'' class, and convolutional layers by the ``\verb|Conv2d|'' class, aligning with PyTorch's naming conventions for parameters and classes.

However, several modifications have been made to accommodate the syntactical constraints and conventions of JavaScript. In JavaScript, functionalities that consume significant time, such as network access, are implemented as asynchronous APIs. Thus, the forward method of models is also implemented as an asynchronous function to enable the usage of asynchronous APIs internally. Consequently, the calling code must use ``\verb|await|'' to await the completion of the processing. Although this introduces some complexity to the implementation, it is essential for future implementations leveraging WebGPU, the next-generation GPU API. Unlike Python, JavaScript does not support defining a class to be callable like a function using the ``\verb|__call__|'' method. Hence, the forward propagation of models is invoked using the ``\verb|model.c|'' method. 
When the model returns multiple tensors, the ``\verb|model.call|'' method is utilized to return an array of tensors.

Furthermore, JavaScript employs garbage collection to automatically release memory that is no longer referenced. However, this mechanism does not function for memory on the GPU. As manual memory release for individual tensors on the GPU is cumbersome, DistML.js utilizes ``\verb|K.tidy|'' function to release memory. ``\verb|K.tidy|'' executes the function passed as an argument and releases tensors generated within that function while tensors or models set as the return value of the function are not released. This idea is inspired by the ``autodestruct'' concept~\cite{hidaka-2017-development} and the implementation in TensorFlow.js~\cite{smilkov-2019}. Unlike PyTorch, tensors in DistML.js are divided into two layers: ``\verb|Tensor|'' (CPUTensor, WebGLTensor) equivalent to NumPy's ndarray~\cite{harris-2020}, which does not contain backpropagation information, and ``\verb|Variable|,'' which wraps these tensors to enable backpropagation. This separation aims to keep the implementation compact for use cases requiring only NumPy-like functionalities as in data preprocessing and traditional machine learning, and to make it easier for third-party developers to add new features.

\section{Experiments}
\label{experiments}

To evaluate the performance of DistML.js, we conducted model training for image classification tasks. The execution environment of DistML.js encompasses edge devices such as PCs and smartphones, where web browsers are available. In this paper, we conducted experiments on two model training methods: training a model using a single edge device and training a model while communicating with multiple edge devices and a central server.

\begin{table*}[t]
  \centering
  \caption{Devices used in the experiments. }
  \begin{tabular}{lll}
    \toprule
    Hardware & OS & Browser \\
    \midrule
    iPhone 13 Mini & iOS & Safari \\
    MacBook Air 2020 (M1) & macOS & Chrome \\
    Desktop PC (AMD Ryzen7 5700X, NVIDIA RTX 4070) & Windows & Edge \\
    \bottomrule
  \end{tabular}
  \label{tab:devices}
\end{table*}

\subsection{Stand-alone training with a single device}
\label{standalone}

In this section, we describe experiments conducted with the aim of quantitatively measuring the training cost of deep learning models on a single device. The model under training is the ResNet18~\cite{he-2016}, a typical CNN model, with 11M trainable parameters. The dataset consists of 10K 32px $\times$ 32px RGB images of from CIFAR10~\cite{krizhevsky-2009}. Upon launching the web application, the dataset is downloaded in bulk, and gradient computations along with model updates are performed on the device. The amount of computation of the forward pass was $7.4\times10^7$ FLOPs. The MomentumSGD optimizer is employed.

\begin{figure}[t]
  \centering
  \includegraphics[width=1\linewidth]{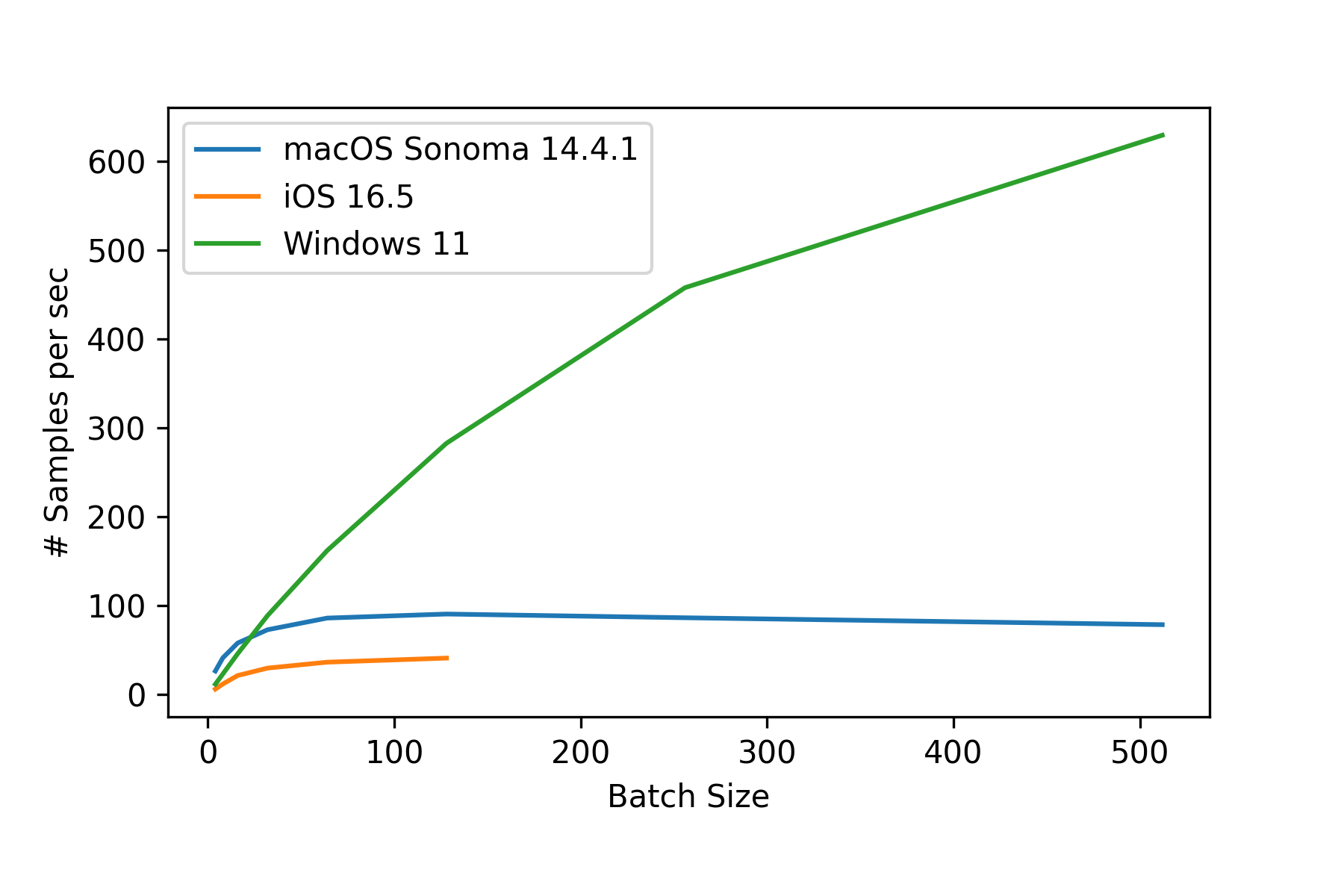}
  \caption{
    The relationship between batch size and processing speed in standalone ResNet-18 training on a single device. The result shows that increasing the batch size leads to improved sample processing speed.}
  \label{fig:standalone}
\end{figure}

To demonstrate that DistML.js works on a variety of hardware and operating systems, we verified its operation on the devices listed in Table~\ref{tab:devices}. To evaluate the performance of model training, the model is trained with batch sizes between 4 to 512 using WebGL backend and the total time taken for one epoch (10K samples) was measured. The sample processing speed per second results are computed and depicted in Fig.~\ref{fig:standalone}. As the batch size increases, the time required per epoch decreases, leading to an increase in the number of samples processed per unit time. While measuring the time required per epoch, gradual changes in processing speed within the same epoch were observed. This is attributed to a slight decrease in processing performance due to heating. The measurement does not consider speed variations within an epoch, indicating the time taken from the start of training to the completion of one epoch. Also, the processing time for validation is not included.

From the above experiments, it is confirmed that processing efficiency improves with increasing batch size. This is likely due to the relative reduction in loop processing overheads by simultaneously performing matrix operations on a large number of samples. Furthermore, for iPhone 13 Mini, it was observed that instability occurs when the batch size is set to 128. This is considered to be due to the limited memory capacity of the GPU usage on a web browser. Based on these results, in the experiments on distributed learning described in Section~\ref{distributed}, the maximum batch size was restricted to 64 empirically.

\subsection{Data distributed learning with multiple devices}
\label{distributed}

In this section, we describe experiments conducted to quantitatively measure the training cost of data-parallel training performed while communicating with multiple devices acting as servers. Data-parallel training is one of the simplest forms of distributed learning and serves as the primary application of DistML.js, which aims to enable volunteer computing without the need for software installation. The system comprises one server and multiple client worker machines. The server, running on a Linux-based machine and implemented in Python, communicates with the workers via HTTP and WebSocket. The workers, running on edge devices, operate web applications using DistML.js.

To implement data-parallel SGD, the following steps are performed at each step of training: (1) The current weights are sent from the server to all client workers. (2) Different mini-batches are sent from the server to each worker as training data. The number of samples sent to one worker is referred to as the local batch size, and the total batch size amoung all workers is referred to as the global batch size. (3) Each worker computes the gradients of the weights. (4) The gradients computed by the workers are sent to the server. (5) The server averages all gradients and updates the model.

Similar to the standalone experiments, the model is ResNet18, the dataset is CIFAR10, and MomentumSGD is used as the optimization method. The training device is an iPhone 13 Mini (iOS 16.5), Safari is used as the web browser, and WebGL is employed as the backend. All workers are connected to a single WLAN access point with a theoretical speed of 1,200~Mbps. The server and access point are wired, with a speed of 1~Gbps.

\begin{figure}[t]
  \centering
  \includegraphics[width=1\linewidth]{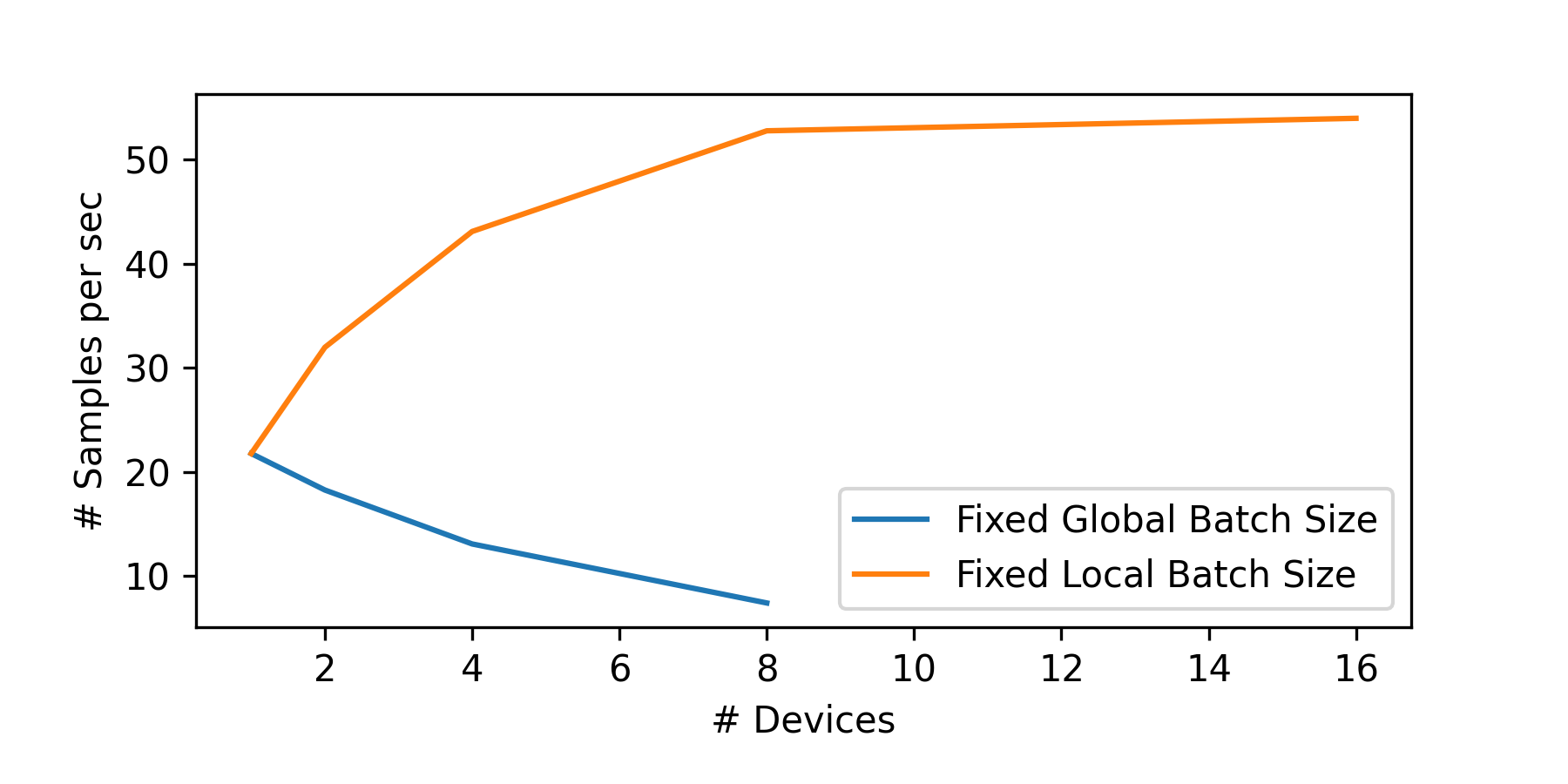}
  \caption{The processing speed in ResNet-18's data parallel distributed learning using multiple devices. Blue and orange plot show the samples processed per second when the global batch size and local batch size are fixed respectively.}
  \label{fig:distributed_fix_global}
\end{figure}

To evaluate the performance of model training in a distributed setting, experiments were conducted under two configurations. One configuration fixed the global batch size at 64 and divided it evenly among the workers. For instance, with four workers, the local batch size would be 16. The other configuration fixed the local batch size at 64 and increased the global batch size proportionally with the number of workers. While the former is expected to yield results consistent with standalone training, the latter may introduce variations. It is known that increasing the learning rate in proportion to the batch size will lead to similar convergence results. In each setting, the total time taken for processing one mini-batch was meaesured, and the samples processed per second was computed.

When the global batch size was fixed, the worker's computational load per iteration was reduced. However, the communication load with the server increased, and much of the mini-batch processing time was dominated by the communication time, which reduced processing speed.
On the other hand, when the local batch size was fixed, the worker's computational load remained unchanged: the communication cost per iteration was the same as when the global batch size is fixed, but the processing speed increased due to the increased number of samples that can be processed in one mini-batch. The processing speed with 16 workers is approximately 2.4 times faster than the speed with a single worker (standalone case).

When the local batch size was 64, the data transmitted per worker per step was 90~MB. The time required for one step with 16 workers was 19 seconds, with the gradient computation taking 1.8 seconds within that time frame. The measured average communication speed was 606~Mbps, indicating that the network bandwidth was the bottleneck. Indeed, as observed in Fig.~\ref{fig:distributed_fix_global}, the batch processing speed per unit time reached a plateau, suggesting that further acceleration is challenging with data parallel SGD.

\section{Relationship to TensorFlow.js}
\label{tfjs}

In this paper, we introduce a novel JavaScript library called DistML.js, which, akin to TensorFlow.js, facilitates the utilization of machine learning models in web browsers. Both libraries share the fundamental concept of enabling machine learning model usage on the web. TensorFlow.js enables the use of TensorFlow functionalities in JavaScript by converting TensorFlow models into JavaScript, thereby enabling the utilization of machine learning models in web browsers. In contrast, DistML.js is designed for training and inference of machine learning models in JavaScript, catering not only to model training within a single device but also supporting distributed training by communicating with servers and other devices over the internet. The design of DistML.js closely resembles the define-by-run API format found in PyTorch. While TensorFlow.js and DistML.js serve different purposes, they complement each other in establishing an environment for utilizing machine learning models in web browsers.
\section{Conclusion}
\label{conclusion}

This paper presents ``DistML.js,'' a JavaScript machine learning library designed to operate within web browsers. Leveraging WebGL backend, DistML.js enables fast computations, allowing model training on edge devices such as PCs and smartphones. Additionally, we explore its application in distributed machine learning, demonstrating the potential for acceleration through model training with multiple edge devices.

In distributed learning experiments, we showed that the training of the model was accelerated by the coordination of 16 workers via web browsers. To achieve efficient learning in more realistic settings, such as when workers are distributed over the internet, methodologies such as performing communication at the epoch level instead of per-step or employing techniques like gradient compression to reduce communication overhead are deemed effective.


\bibliography{example_paper}
\bibliographystyle{mlsys2024}

\end{document}